\documentclass[10pt,twocolumn,letterpaper]{article}

\usepackage{cvpr}
\usepackage{times}
\usepackage{epsfig}
\usepackage{graphicx}
\usepackage{amsmath,bm}
\usepackage{amssymb}
\usepackage{multirow}
\usepackage{color}

\cvprfinalcopy 

\def\cvprPaperID{2562} 

\ifcvprfinal\fi
\begin{document}

\title{Signal-to-Noise Ratio: A Robust Distance Metric for Deep Metric Learning}
\author{Tongtong Yuan\textsuperscript{1}, Weihong Deng\textsuperscript{1}\thanks{Corresponding author}, Jian Tang\textsuperscript{2,3}, Yinan Tang\textsuperscript{1}, Binghui Chen\textsuperscript{1}\\
\textsuperscript{1}Beijing University of Posts and Telecommunications, Beijing, China\\
\textsuperscript{2}AI Labs, Didi Chuxing, Beijing, China\\
\textsuperscript{3}Department of Electrical Engineering and Computer Science, Syracuse University, Syracuse, NY USA\\
{\tt\small \{yuantt, whdeng, tn513, chenbinghui\}@bupt.edu.cn, tangjian@didiglobal.com}}


\maketitle
\thispagestyle{empty}

\begin{abstract}
Deep metric learning, which learns discriminative features to process image clustering and retrieval tasks, has attracted extensive attention in recent years. A number of deep metric learning methods, which ensure that similar examples are mapped close to each other and dissimilar examples are mapped farther apart, have been proposed to construct effective structures for loss functions and have shown promising results. In this paper, different from the approaches on learning the loss structures, we propose a robust SNR distance metric based on Signal-to-Noise Ratio (SNR) for measuring the similarity of image pairs for deep metric learning. By exploring the properties of our SNR distance metric from the view of geometry space and statistical theory, we analyze the properties of our metric and show that it can preserve the semantic similarity between image pairs, which well justify its suitability for deep metric learning. Compared with Euclidean distance metric, our SNR distance metric can further jointly reduce the intra-class distances and enlarge the inter-class distances for learned features. Leveraging our SNR distance metric, we propose Deep SNR-based Metric Learning (DSML) to generate discriminative feature embeddings. By extensive experiments on three widely adopted benchmarks, including CARS196, CUB200-2011 and CIFAR10, our DSML has shown its superiority over other state-of-the-art methods. Additionally, we extend our SNR distance metric to deep hashing learning, and conduct experiments on two benchmarks, including CIFAR10 and NUS-WIDE, to demonstrate the effectiveness and generality of our SNR distance metric.
\end{abstract}

\section{Introduction}
Recent years have witnessed the extensive research on metric learning, which aims at learning semantic distance and embeddings such that similar examples are mapped to nearby points on a manifold and dissimilar examples are mapped apart from each other~\cite{oh2016deep,sohn2016improved,ustinova2016learning,xing2003distance}. Compared to conventional distance metric learning, deep metric learning learns a nonlinear embedding of the data using deep neural networks, and it has shown significant benefits by exploring more loss structures. With the development of these learning techniques, deep metric learning has been widely applied to the tasks of face recognition~\cite{taigman2014deepface,sun2014deep}, image clustering and retrieval~\cite{wang2014learning,oh2016deep}.

Deep metric learning has made remarkable successes in generating discriminative features. To improve the performance of learned features, many learning methods have explored the structures in the objective functions, such as contrastive loss~\cite{hadsell2006dimensionality}, triplet loss~\cite{schroff2015facenet,weinberger2006distance}, lifted structured embedding~\cite{oh2016deep}, N-pair Loss method~\cite{sohn2016improved},~\etc. These deep metric learning methods can be categorized as \textit{structure-learning} methods, which focus on constructing more effective structures for objective functions by making use of training batches or increasing negative examples. However, most structure-learning methods simply take the Euclidean distance as the semantic distance metric and ignore that the distance metric is playing a nonnegligible role in deep metric learning. Different from structure-learning, some metric learning methods~\cite{weinberger2009distance,davis2007information} employ new distance metrics to metric learning. For example, Weinberger~\etal have proposed a distance metric for k-nearest neighbor (kNN) classification in metric learning, i.e, Mahalanobis distance~\cite{weinberger2009distance}, which shows that the performance of metric learning algorithms also depends on the distance metric. Contrary to structure-learning methods, these methods exploring a new distance metric can be categorized as \textit{distance-learning} methods. Compared to the structure-learning methods, designing a good distance metric for measuring the semantic similarity may make a more significant impact on learning discriminative embeddings. Therefore, we focus on designing of a novel and effective distance metric.

Measuring similarities between pairs of examples is critical for metric learning. The most well-known distance metric is Euclidean distance, which has been widely used in learning discriminative embeddings. However, Euclidean distance metric only measures the distance between paired examples in $n$-dimensional space, lacking the abilities to preserve the correlation and improve the robustness of the pairs. Therefore, we devise a new distance metric by leveraging a concept defined in signal processing, \ie Signal-to-Noise Ratio~(SNR), as a similarity measurement in deep metric learning. Generally, SNR in signal processing is used to measure the level of a desired signal to the level of noise, and a larger SNR value means a higher signal quality. For similarity measurement in deep metric learning, a pair of learned features $\bm{x}$ and $\bm{y}$ can be given as $\bm{y}=\bm{x}+\bm{n}$, where $\bm{n}$ can be treated as a noise. Then, the SNR is the ratio of the feature variance and the noise variance. Based on the definition of SNR in deep metric learning, we find that SNR is promising to be formulated as a distance metric for measuring the differences between paired features.


In this paper, based on the properties in SNR, we propose an SNR distance metric to replace Euclidean distance metric for deep metric learning. In the aspect of space analysis and theoretical demonstration, we explain the advantages of SNR distance over Euclidean distance. Different from Euclidean distance, SNR distance is a more robust distance metric, which can further jointly reduce the intra-class distances and enlarge the inter-class distances for the learned features, and preserve the correlations of the features. Moreover, we propose a \textbf{D}eep \textbf{S}NR-based \textbf{M}etric \textbf{L}earning~(DSML) method, which uses SNR distance metric as similarity measurement for generating more discriminative features. To show the generality of our SNR-based metric, we also extend our approach to hashing retrieval learning.

Our main contributions can be summarized as follows. (1) To the best of our knowledge, this is the first work that employs SNR to build the distance metric in deep metric learning. By analyzing the properties of the SNR distance metric, we find that it has better performance than Euclidean distance and can be widely used in deep metric learning. (2) We show how to integrate our SNR distance metric into the popular learning frameworks, and propose the corresponding objective function in our DSML. (3) We make extensive experiments on three widely-used benchmarks about image clustering and retrieval tasks, and the results demonstrate the superiority of our deep SNR-based metric learning approach over state-of-the-art methods. (4) We extend our SNR-based metric distance to deep hashing learning and obtain promising experiment results.

\section{Related Work}
\subsection{Metric Learning}
Metric learning methods, which have been widely applied to image retrieval, clustering and recognition tasks, have attracted much attention. With the development of deep neural networks, deep metric learning methods~\cite{cui2016fine,paisitkriangkrai2015learning,liao2015person,huang2016local} have shown promising performance on the complex computer vision tasks. To distinguish the innovations of different deep metric learning methods, we roughly divide these approaches into structure-learning and distance-learning methods, and introduce these works briefly. Related to our work, we also introduce deep hashing methods based on the famous metric learning structures.
\subsubsection{Structure-Learning Methods}
The most well-known structure-learning approach is contrastive embedding, which is proposed by Hadsell~\etal~\cite{hadsell2006dimensionality}. The main idea of contrastive loss~\cite{hadsell2006dimensionality} is that similar examples should be mapped to nearby points on a manifold and dissimilar examples should be mapped apart from each other. This idea have established the foundation of the objective functions in deep metric learning. Following this work, the subsequent structure-learning methods have proposed various loss functions with different structures. For example, triplet loss~\cite{schroff2015facenet,weinberger2006distance} is composed of triplets, and each triplet is consisted of a anchor example, a positive example and a negative example. The triplet loss encourages the positive distance to be smaller than the negative distance with a margin. Lifted structured loss~\cite{oh2016deep} lifts the vector of pairwise distances within the batch to the matrix of pairwise distances. N-pair loss~\cite{sohn2016improved} generalizes triplet loss by allowing joint comparison among more than one negative examples, which means a feature pair is composed of samples from the same labels and other pairs in the mini-batch have different labels. ALMN \cite{chen2018almn} proposes to optimize an adaptive large margin objective via the generated virtual points instead of mining hard-samples. Besides these works, several works~\cite{schroff2015facenet,shrivastava2016training} try to mine hard negative data on the basis of triple loss, and they can been seen as enhanced structure-learning methods. Different from these structure-learning methods, our work aims to design a new distance metric for deep metric learning. Because most structure-learning methods use the Euclidean distance as their similarity measurement (inner product in N-pair loss can be regarded as a similar Euclidean measurement), they can provide the baselines for our work.

\subsubsection{Distance-Learning Methods}
Different from structure-learning approaches, the distance-learning method, which explores a superior distance metric, is also promising to improve the performance of deep metric learning. In traditional metric learning~\cite{schultz2004learning,shalev2004online}, some distance-learning methods have been proposed by using Mahalanobis distance to measure the similarities of samples. For instance, Globerson~\etal~\cite{globerson2006metric} presented an algorithm to learn Mahalanobis distance in classification tasks. Weinberger~\etal~\cite{weinberger2009distance} showed how to learn a Mahalanobis distance metric for kNN classification from labeled examples. Davis~\etal~\cite{davis2007information} presented an information-theoretic approach to learning a Mahalanobis distance function. In deep metric learning, we noticed that in order to learn better features, Wang \etal proposed a distance-learning method to constrain the angle at the negative point of triplet triangles~\cite{wang2017deep}. Moreover, Chen \etal \cite{chen2019energy} introduce energy confusion metric to improve the generalization of the learned deep metric. Chen \etal \cite{chen2019hybrid} propose the hybrid-attention based decoupled metric learning framework for learning discriminative and robust deep metric. However, the angle measurement for triangles has limitations when measuring the distance of two points, and it cannot be regarded as a general distance metric. In this paper, we propose a general distance-learning method, which uses SNR-based metric for measuring the similarity of image pairs in deep metric learning.

\subsection{Hashing Learning}
Similar to deep metric learning, deep hashing aims to learn a discriminative embedding to preserve the consistency with semantic similarity in binary features. Recently, many deep hashing methods~\cite{xu2018sketchmate,lin2016learning,xia2014supervised,Yuan2018Supervised,liu2017deep,venkateswara2017deep,shen2018unsupervised,yuan2019unsupervised} have been proposed to learn compact binary codes and retrieve the similar images in Hamming space. Benefiting from metric learning methods, some deep hashing methods ~\cite{liu2016deep,li2015feature, wang2016deep} are established on contrastive embedding or triplet embedding. In this paper, in order to extend the application of our SNR-based metric and verify the generality of the metric, we also propose a deep SNR-based hashing learning method, which aims to generate similarity-preserving binary codes by training the convolutional neural networks with our SNR metric based loss layer.

\section{Proposed Approach}
Pair-wise distances in features are usually measured by Euclidean distance metric, which has been rarely changed~\cite{wang2017deep}. However, designing a good distance metric for measuring the similarity between images is significant for improving the performance of deep metric learning. Therefore, we propose a new SNR-based metric for deep metric learning.
\subsection{SNR-based Metric}

\textbf{Definition:}
In deep metric learning, given two images $\bm{x}_i$ and $\bm{x}_j$, the learned features can be denoted as $\bm{h}_i=f(\theta;\bm{x}_i)$ and $\bm{h}_j=f(\theta;\bm{x}_j)$, where $f$ is the metric learning function and $\theta$ denotes the learned parameters. Given a pair of features $\bm{h}_i$ and $\bm{h}_j$, where the anchor feature is $\bm{h}_i$ and the compared feature is $\bm{h}_j$. We denote the anchor feature $\bm{h}_i$ as signal, and the compared feature $\bm{h}_j$ as noisy signal, then the noise $\bm{n}_{ij}$ in $\bm{h}_i$ and $\bm{h}_j$ can be formulated as $\bm{n}_{ij}=\bm{h}_j-\bm{h}_i$.

In statistical theory, a standard definition of SNR is the ratio of signal variance to noise variance~\cite{dicker2014variance}, so we define the SNR between the anchor feature $\bm{h}_i$ and the compared feature $\bm{h}_j$ as:
\begin{small}
\vspace{-1em}
\begin{equation}\label{e1}
\text{SNR}_{i,j}=\frac{var(\bm{h}_i)}{var(\bm{h}_j-\bm{h}_i)}=\frac{var(\bm{h}_i)}{var(\bm{n}_{ij})},
\end{equation}
\end{small}
where $var(\bm{a})=\frac{\sum_{i=1}^n(a_i - \mu)^2}{n}$ denotes the variance of $\bm{a}$, and $\mu$ is the mean value of $\bm{a}$. If $\mu=0$, $var(\bm{a})=\frac{\sum_{i=1}^n(a_i)^2}{n}$.

The variance in information theory reflects the informativeness. More explicitly, the signal variance measures the useful information, while the noise variance measures the useless information. Therefore, increasing $\text{SNR}_{i,j}$ can improve the ratio of useful information to useless information, which reflects the compared feature can be more similar to the anchor feature. On the contrary, decreasing $\text{SNR}_{i,j}$ can increase the proportion of noise information, leading to more difference in the two features. Therefore, the values of $\text{SNR}_{i,j}$ can be used to measure the difference in a pair of features reasonably, which is an essential to construct a distance metric in metric learning.

\textbf{SNR distance metric:}
In deep metric learning, the constraint of most loss functions based on Euclidean distance metric is that similar examples should have short distances in features while dissimilar examples should have large distances in features. According to the constraint, we design a new distance metric as similarity measurement for deep metric learning. On the basis of the definition of SNR, we propose our SNR distance metric. The SNR distance $d_S$ in a pair of features $\bm{h}_i$ and $\bm{h}_j$ is defined as:
\begin{small}
\vspace{-1em}
\begin{equation}\label{snr2}
  d_S(\bm{h}_i,\bm{h}_j)=\frac{1}{\text{SNR}_{ij}}=\frac{var(\bm{n}_{ij})}{var(\bm{h}_i)}.
\end{equation}
\end{small}
Notably, the commutative property $(d_E(\bm{h}_i,\bm{h}_j)=d_E(\bm{h}_j,\bm{h}_i))$ in Euclidean distance $d_E$ is inapplicable in our SNR distance. Because the values of $d_S(\bm{h}_i,\bm{h}_j)$ and $d_S(\bm{h}_j,\bm{h}_i)$ are usually not equal, our SNR distance is sensitive to which one is the anchor feature in a pair.

\begin{figure}[t]
\centering
\includegraphics[width=0.35\textwidth]{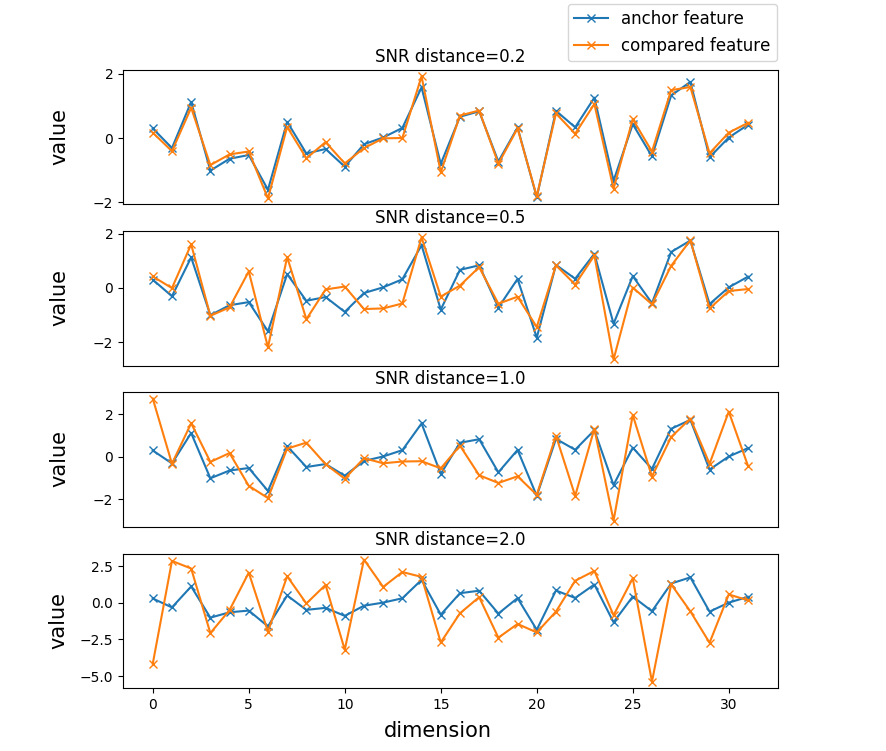}
\caption{\label{ff1} The curves show the comparisons of 32-dimensional synthetic anchor features and the compared features under different SNR distances.}
\vspace{-1em}
\end{figure}

To show how SNR distance reflects the differences in a pair of features, we synthesize a 32-dimensional Gaussian data with $N\sim\{0,1\}$ as anchor feature, and a series of Gaussian noises with $N\sim \{0,\sigma^2\}$, where $\sigma^2=\{0.2,0.5,1.0,2.0\}$. The compared feature is synthesized by adding the noise data to the anchor feature, then the SNR distance $d_S$ of the anchor feature and compared feature is $\{0.2,0.5,1.0,2.0\}$. As shown in Figure~\ref{ff1}, the longer SNR distance reflects that the difference between the anchor feature and the compared feature is larger. Therefore, the SNR distance applied to the loss functions can have a similar property with Euclidean distance (\ie, similar image pairs are supposed to have a short SNR distance in features, while dissimilar image pairs should have a large SNR distance in features). As a result, we can use the SNR distance metric as the similarity measurement to replace the Euclidean distance metric in deep metric learning.

\textbf{Superiority analysis:}
To indicate the superiority of SNR distance to Euclidean distance, we compare these two metrics from the view of geometry space and statistical theory.

The Euclidean distance of two points $\bm{a}$ and $\bm{b}$ is defined as:
\vspace{-1em}
\begin{small}
\begin{equation}\label{e2}
d_E(\bm{a},\bm{b})=\sqrt{\sum_{i=1}^n(a_i-b_i)^2}.
\end{equation}
\end{small}
\vspace{-0.5em}

For SNR distance, according to Equations~(\ref{snr2}) and~(\ref{e2}), we can derive that if the features follow zero-mean distributions:
\vspace{-1em}
\begin{small}
\begin{equation}\label{e3}
\begin{split}
  d_S(\bm{h}_j,\bm{h}_i)&=\frac{var(\bm{n}_{ij})}{var(\bm{h}_i)}=\frac{\sum_{m=1}^M(h_{im}-h_{jm})^2}{\sum_{m=1}^M(h_{im})^2}\\
  &=\frac{{d_E({\bm{h}_i,\bm{h}_j})}^2}{{d_E({\bm{h}_i})}^2},
\end{split}
\end{equation}
\end{small}
\vspace{-0.5em}
\begin{figure}[t]
\centering
\includegraphics[width=0.5\textwidth]{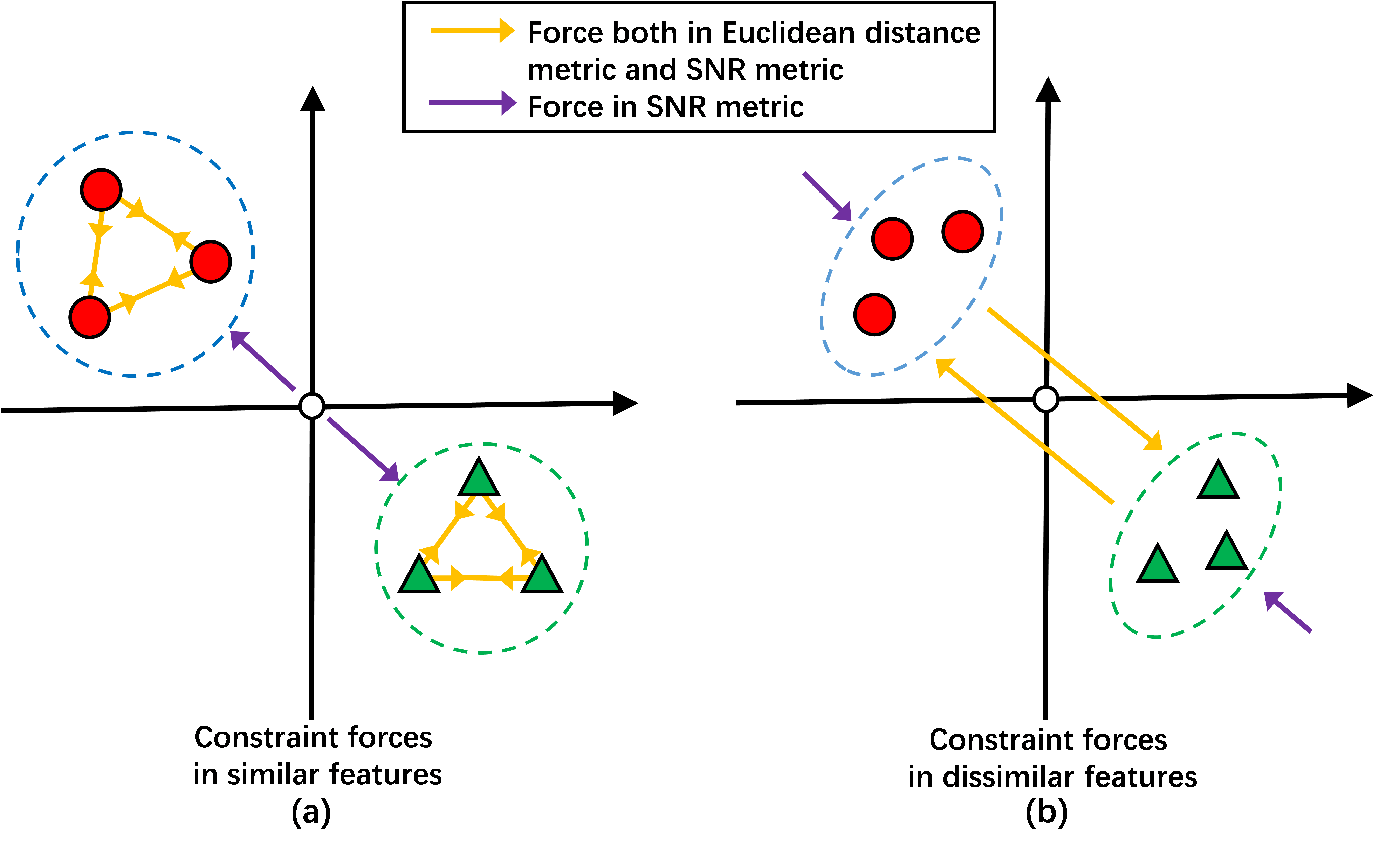}
\caption{\label{ff0} This example shows how SNR distance metric and Euclidean metric affect the features in Euclidean space. The constraints for preserving the semantic similarity are described as repulsive forces and attractive forces. The arrowed lines represent forces, where the purple lines denote the forces only from the SNR distance metric, and orange lines are the forces shared by Euclidean distance and SNR distance. As shown in (a), for similar images, minimizing Euclidean distance can only reduce the distances between the intra-class examples. Because our SNR distance takes into account the Euclidean distance from the feature to the origin, minimizing SNR distance can also enlarge the inter-class Euclidean distances.
As shown in (b), for dissimilar samples, the Euclidean distance of the inter-class examples should be increased. Different from the constraint force of Euclidean metric, the constraint forces caused by increasing SNR distance (\ie orange lines and purple lines) can collaborate to make each cluster more compact, leading to the smaller intra-class distances.}
\vspace{-1em}
\end{figure}

where $d_E({\bm{h}_i})$ denotes the Euclidean distance from $\bm{h}_i$ to the origin $O$, and $M$ is the dimension of learned features $\bm{h}$. As shown in (\ref{e3}), besides the Euclidean distance of the paired features, the SNR distance also takes into account the Euclidean distance from the feature to the origin.

In order to preserve the semantic similarity, the loss functions with Euclidean distance metric constrain that the Euclidean distances in feature pairs with the same labels should be reduced, while the Euclidean distances in feature pairs with the different labels should be increased. Different from Euclidean distance metric, the loss functions with SNR distance metric can make an additional constraint on the Euclidean distance from origin to the features. As shown in Figure~\ref{ff0}, compared to Euclidean distance metric which only measures the Euclidean distances of feature pairs, our SNR distance can not only provide the constraints in Euclidean distances, but also give an additional constraint to enlarge the inter-class distances when dealing with similar pairs, and to reduce the intra-class distances when dealing with dissimilar pairs. As a result, in deep metric learning, our SNR distance metric is more powerful to increase the discrimination and robustness of feature pairs.

We also explore the relationship between SNR distance and the correlation coefficient of paired features to further show the superiority to Euclidean distance,
If the mean of each feature is zero, and the noise is independent to the signal feature, the correlation coefficient $corr(\cdot,\cdot)$ in paired features can be computed via the statistical theory as follows:
\vspace{-1em}
\begin{small}
\begin{equation}\label{rr13}
\begin{split}
  &corr(\bm{h_i},\bm{h_j})=\frac{cov(\bm{h_i},\bm{h_j})}{\sqrt{var(\bm{h_i})var(\bm{h_j})}}=\frac{E(\bm{h_ih_j})}{\sqrt{var(\bm{h_i})var(\bm{h_j})}}\\
  &=\frac{E(\bm{h_i}(\bm{h_i}+\bm{n_{ij}}))}{\sqrt{var(\bm{h_i})var(\bm{h_i}+\bm{n_{ij}})}}=\frac{E(\bm{h_i}^2)}{\sqrt{var(\bm{h_i})var(\bm{h_i}+\bm{n_{ij}})}}\\
  &=\frac{var(\bm{h_i})}{\sqrt{var(\bm{h_i})^2+var(\bm{h_i})var(\bm{n_{ij}})}}=\frac{1}{\sqrt{1+\frac{var(\bm{n_{ij}})}{var(\bm{h_i})}}}\\
  &=\frac{1}{\sqrt{1+\frac{1}{\text{SNR}_{ij}}}}=\frac{1}{\sqrt{1+d_S(\bm{h}_j,\bm{h}_i)}}.
\end{split}
\end{equation}
\end{small}
\vspace{-1em}

According to (\ref{rr13}), the correlation coefficient of the paired features is an decreasing function of their SNR distance. Increasing the SNR distance will reduce the correlation in dissimilar features, and reducing the SNR distance will increase the correlation in similar pairs. Therefore, by using the SNR distance instead of Euclidean distance, deep metric learning can jointly preserve the semantic similarity and the correlations in learned features.

\subsection{Deep SNR-based Metric Learning}
Because of the superiority of SNR distance metric, the SNR distance can provide a more effective similarity measurement compared with the Euclidean distance. Besides, the SNR distance can be generally applied to various objective functions of deep metric learning. In order to realize deep SNR-based metric learning (DSML), we select four attractive deep metric learning structures, including contrastive loss~\cite{hadsell2006dimensionality}, triplet loss~\cite{schroff2015facenet,weinberger2006distance}, lifted structured loss~\cite{oh2016deep}, and N-pair loss~\cite{sohn2016improved}, to construct our SNR-based objective functions.

In DSML, we denote the learned features as $\bm{h}_i\in (\bm{h_1},\cdots,\bm{h_N})$. For an anchor feature $\bm{h}_i$, the positive feature is $\bm{h}_i^+$, and the negative one is denoted as $\bm{h}_i^-$. Based on SNR distance metric, the distance of two features $\bm{h}_i$,$\bm{h}_j$ in our DSML functions can be represented as:

\vspace{-0.7em}
\begin{small}
\begin{equation}
{d_S}_{ij}=d_S(\bm{h}_i,\bm{h}_j)=\frac{1}{\text{SNR}}=\frac{var(\bm{h}_i-\bm{h}_j)}{var(\bm{h}_i)}.
\end{equation}
\end{small}
\vspace{-1em}

We use a regularization $\lambda L_r$ to constrain that the features have zero-mean distributions, and the regularization is defined as:
\vspace{-1em}
\begin{small}
\begin{equation}\label{e11}
  L_r= \lambda\frac{1}{N}\sum_{i\in N} \mid\sum_{m=1}^M h_{im}\mid,
\end{equation}
\end{small}
where $\lambda$ is a hyper-parameter with a small value.

Combined with the four learning structures, the SNR-based objective functions of our DSML are detailed in the following.

\textbf{DSML(cont):}
For SNR-based contrastive embedding, our DSML objective function is:
\vspace{-1em}
\begin{small}
\begin{equation}\label{cn2}
  J=\sum^{N_i}_{i=1} d_S(\bm{h}_i,\bm{h}_i^+)+\sum^{N_j}_{j=1} {[\alpha-d_S(\bm{h}_j,\bm{h}^-_j)]}_+ +\lambda L_r,
\end{equation}
\end{small}
where $N_i$ and $N_j$ respectively represent the numbers of positive and negative pairs, $\alpha$ denotes the margin to constrain the negative pairs, and $[\cdot]_+$ denotes the function $\text{max}(0,\cdot)$.


\textbf{DSML(tri):}
For SNR-based triplet embedding, the objective function is defined as:
\vspace{-1em}
\begin{small}
\begin{equation}\label{et2}
  J=\sum^{N}_{i=1} [d_S(\bm{h}_i,\bm{h}_i^+)-d_S(\bm{h}_i,\bm{h}_i^-)+\alpha]_+ +\lambda L_r,
\end{equation}
\end{small}
\vspace{-1em}

which constrains that the positive SNR distance should be smaller than the negative SNR distance with a margin $\alpha$. In triplet embedding learning, we generate all the valid triplets and average the loss over the positive ones.


\textbf{DSML(lifted):}
For SNR-based lifted loss function, we deploy the SNR distance ${d_S}_{ij}$ as follows:
\vspace{-1em}
\begin{small}
\begin{equation}
\begin{split}
  &\ \ J=\frac{1}{2N_i}\sum_{(i,j)\in \widehat{P}} \max(0, J_{i,j})+\lambda L_r,\\
  &J_{i,j}= \max(\max_{(i,k)\in \widehat{N} } \alpha-\beta {d_S}_{ik}, \max_{(j,l)\in \widehat{N} } \alpha-\beta {d_S}_{jl})+\beta {d_S}_{ij} ,
\end{split}
\end{equation}
\end{small}

where $\widehat{P}$ and $\widehat{N}$ denote positive pairs and negative pairs, $\alpha$ denotes margin, and $\beta$ is a hyper-parameter to ensure the convergence of loss.

\begin{table*}[t]
\caption{\label{tab1}Results on CARS196 with Alexnet.}
\centering
\begin{tabular}{|c|ccc|ccc|ccc|ccc|}
\hline
Tasks&\multicolumn{6}{|c|}{Image Clustering}&\multicolumn{6}{|c|}{Image Retrieval}\\
\hline
 score (\%) & \multicolumn{3}{|c|}{F1}&\multicolumn{3}{|c|}{NMI}&\multicolumn{3}{|c|}{Recall@1}&\multicolumn{3}{|c|}{Recall@2}\\
\hline
 embedding size&16&32&64& 16&32&64&16&32&64&16&32&64\\
\hline
contrastive&9.2&10.6&11.0 & 31.5& 34.4&33.3&8.9& 14.0& 16.3 & 10.3 &	16.1 &	18.4 \\

DSML(cont)& 12.9&11.9& 11.8&39.9& 37.0&36.1&15.1 & 16.5 & 18.0 &17.5& 	18.6&201 \\
\hline
triplet&19.4&16.9& 15.4& 50.9& 47.9& 46.8&24.8&  20.6 &  19.5 &28.2& 23.5& 22.1\\
DSML(tri)& 25.6&\textbf{33.1}&\textbf{34.4} &52.5& \textbf{56.8}&\textbf{57.4}& \textbf{38.5} &\textbf{46.3} & \textbf{49.1} &\textbf{42.0} &\textbf{49.8} &\textbf{52.4}\\
 \hline
lifted& 27.1& 29.0 & 28.1 & 53.1&54.4&53.9& 37.2&39.1&40.6 &41.2 &42.9 &44.3\\
DSML(lifted)& 30.2& 32.1& 33.6 &54.1& 55.6 & 56.7&35.3&40.3&43.8&38.9 &	44.0 &47.5\\
\hline
N-pair& 26.9&29.9&29.5&51.8&53.5&53.6& 32.9& 36.3& 38.3&36.7 &	39.8 &	42.1\\
DSML(N-pair)&\textbf{30.7}& \textbf{33.1}& 32.7 &\textbf{54.5}& 54.4& 56.4&37.8&40.4&44.9& 39.8 &	44.5 &	48.6\\
\hline
\end{tabular}
\end{table*}

\begin{table*}[t]
\caption{\label{tab2}Results on CUB200-2011 with Alexnet.}
\centering
\begin{tabular}{|c|ccc|ccc|ccc|ccc|}
\hline
Tasks&\multicolumn{6}{|c|}{Image Clustering}&\multicolumn{6}{|c|}{Image Retrieval}\\
\hline
 score(\%)&\multicolumn{3}{|c|}{F1}&\multicolumn{3}{|c|}{NMI}&\multicolumn{3}{|c|}{Recall@1}&\multicolumn{3}{|c|}{Recall@2}\\
\hline
 embedding size&16&32&64& 16&32&64&16&32&64&16&32&64\\
\hline
contrastive&14.6&18.7&19.3 & 41.6& 46.6&47.4&15.8& 25.7& 29.7 & 18.0 &28.6 &32.7\\
DSML(cont) & 19.6&19.7& 22.7&47.5& 47.8&50.5&22.2 & 27.2 & 33.1& 25.3 &30.6 &	36.4 \\
\hline
triplet&23.6&22.1& 21.7& 56.5& 55.6& 55.3&33.9&  32.8 &  32.6&37.8 &36.4 &	35.6\\

DSML(tri) & 36.1&39.0&40.3 &63.0& 64.0&\textbf{65.6}& 45.7 & \textbf{49.8} & \textbf{51.6}& 49.3 &	\textbf{53.5} &\textbf{54.9}\\
 \hline
lifted & 36.0& 36.5 & 37.2 & 60.9&61.1&61.4& 43.2&44.5&46.8&46.4 &	47.8 &50.4\\

DSML(lifted)& \textbf{41.3}& \textbf{43.9}& \textbf{45.8} &\textbf{63.5}& 6\textbf{4.5} & 65.4&\textbf{46.0}&48.8&51.0&\textbf{49.4} &51.9& 54.4\\
\hline
N-pair& 34.7&35.7&37.6&59.6&60.0&61.5& 39.9& 40.7& 43.1&43.3 &	44.4& 46.9\\

DSML(N-pair) &37.6& 38.1& 40.5&62.4& 61.9&63.1&42.3&46.2&48.5&48.6 &	49.7 &51.9 \\
\hline
\end{tabular}
\end{table*}

\begin{figure*}[t]
\centering
\includegraphics[width=0.8\textwidth]{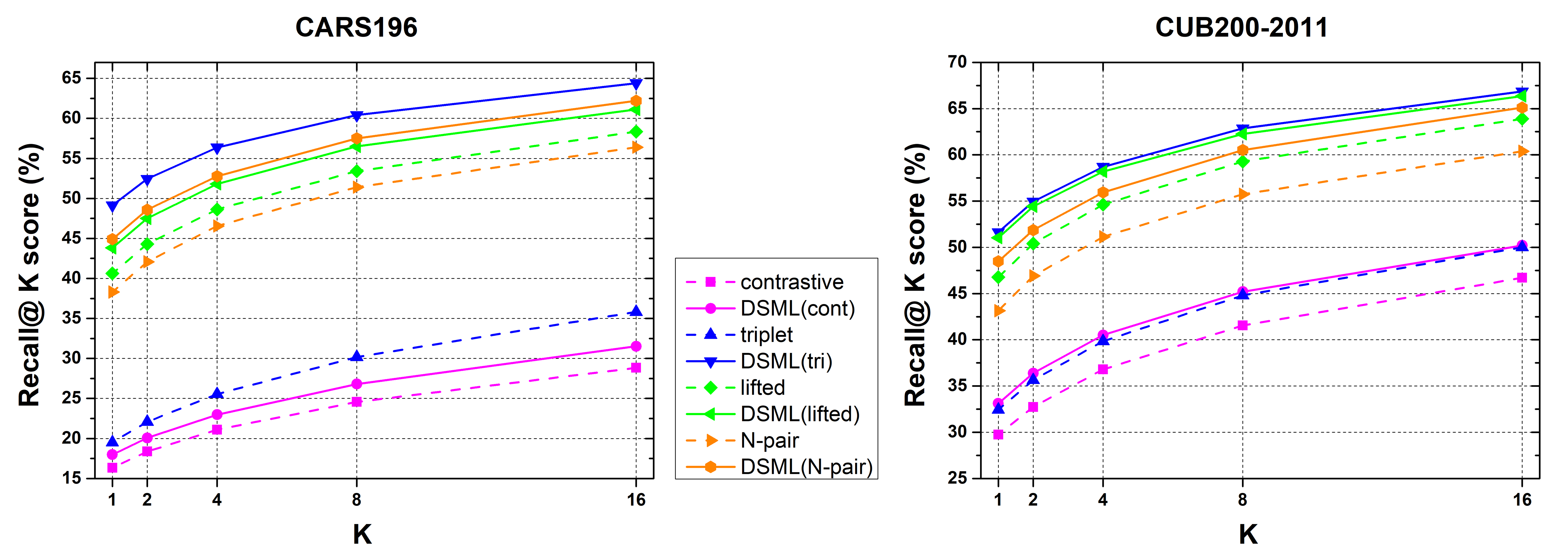}
\caption{\label{ff2} Recall@K curves on CARS196 and CUB200-2011 at embedding size of 64. Dashed lines denote Euclidean-based methods and solid lines   represent SNR-based methods.}
\end{figure*}

%

\textbf{DSML(N-pair):}
In the original N-pair loss, each tuplet $T_i$ is composed of $\{x_i, x^+_1 , x^+_2 ,\cdots, x^+_N\}$, where $x_i$ is the query for $T_i$, $x^+_i$ is the positive example, and $x^+_ j$ ($j\neq i$) are the negative examples. The N-pair loss function is constructed by similarity rather than distance, and the similarity is measured by the inner product $S_{ij}=\bm{h}_i^T\bm{h}_j$, which cannot be directly replaced by our SNR distance metric. Therefore, in our DSML(N-pair), we construct a SNR-based similarity to adapt our SNR-based metric to N-pair learning framework. The similarity $S_{ij}$ of $h_i$ and $h_j$ for DSML(N-pair) is:
\vspace{-1em}
\begin{small}
\begin{equation}
  S_{ij}=\frac{1}{{{d_S}_{ij}}^2}=\text{SNR}_{ij}^2= \frac{var(\bm{h}_i)^2}{var(\bm{h}_i-\bm{h}_j)^2}.
\end{equation}
\end{small}
\vspace{-1em}

 Then, the objective function of DSML(N-pair) is:
\vspace{-1em}
\begin{small}
\begin{equation}\label{np1}
  J=\frac{1}{N} \sum_{i=1}^N log(1+\sum_{j\neq i}exp(S_{ij^+} - S_{ii^+}))+\lambda L_r
\end{equation}
\end{small}
\vspace{-1em}

In summary, the objective functions defined in our DSML are easily to be formulated with the guide of the state-of-the-art methods in deep metric learning, which implies that our SNR-based metric have a good generality, and it is promising to be widely applied in deep embedding learning.

\subsection{Deep SNR-based Hashing Learning}
Hashing learning methods aim to generate discriminative binary codes for image samples, where the binary codes of similar images have short Hamming distances, and the binary codes of dissimilar images have long Hamming distances.
To indicate the generality of our SNR-based metric, we deploy our SNR distance metric to deep hashing learning.

By using SNR-based contrastive loss (\ref{cn2}) as the objective function, we proposed Deep SNR-based Hashing method (DSNRH). The main difference between the deep metric learning and the deep hashing learning is that the learned embeddings need to be quantized to binary features in hashing. Thus, in our DSNRH, after learning the features $\bm{h}$, we use the sign function $\bm{B}=sign(\bm{h})$ to generate binary codes for Hamming space retrieval, where the binary codes $\bm{B}$ is consisted of $M$-bit binary codes. Similar to the existing hashing learning methods~\cite{li2015feature,wang2016deep}, the similarity labels are given as: if two images $i$ and $j$ share at least one label, they are similar, otherwise they are dissimilar.

\vspace{-1em}
\begin{table*}[t]
\caption{\label{tab3}Retrieval results on CIFAR10 with AlexNet.}
\centering
\begin{tabular}{|c|ccc|ccc|ccc|ccc|}
\hline
 &\multicolumn{6}{|c|}{Euclidean Ranking}&\multicolumn{6}{|c|}{Hamming Ranking}\\
\hline
 score (\%) & \multicolumn{3}{|c|}{MAP@59000}&\multicolumn{3}{|c|}{F1@5000}&\multicolumn{3}{|c|}{MAP@59000 }&\multicolumn{3}{|c|}{F1@5000 }\\
\hline

 embedding size&16&32&64& 16&32&64&16&32&64& 16&32&64\\
\hline
contrastive&75.5&73.4&69.3 & 69.1& 67.2&61.4&65.5& 66.9& 61.8&61.2 & 62.2&56.9 \\

DSML(cont) & \textbf{80.0}&\textbf{79.8}& \textbf{79.0}&\textbf{72.9}& \textbf{72.7}&\textbf{72.1}&\textbf{73.7} &\textbf{76.6} &\textbf{76.9} &\textbf{70.0} & \textbf{72.2}& \textbf{71.4} \\
\hline
triplet&75.9&77.3& 75.8& 70.7& 71.2& 70.3&71.9&73.7 & 74.3& 67.3& 70.2& 69.8\\

DSML(tri) & 78.4&78.3&77.4 &72.4& 72.5&71.6&73.4& 74.5 & 75.3& 69.9& 70.8 & 70.8\\
\hline
lifted & 63.7& 54.6& 55.5 &60.6& 52.0&52.0&60.3&52.1&53.9&54.9&50.0&50.8\\

DSML(lifted)& 78.1& 76.2& 76.7 &73.5& 71.1&71.8&66.9&74.3&70.7&58.1&70.5&67.1\\
\hline
N-pair& 53.5&51.1&39.5&49.5&47.5&37.8& 48.4& 48.9& 38.6& 45.9& 46.4&37.3\\

DSML(N-pair)&62.1&64.1&56.6&57.1&58.8&52.1& 55.2& 62.0& 53.6& 50.2& 57.3&49.6\\
\hline
\end{tabular}
\end{table*}
\section{Experiments}
We mainly conduct experiments on deep metric learning, and also compare our DSNRH with some state-of-the-art deep hashing methods.

\subsection{Experiments on Deep Metric Learning }
\subsubsection{Datasets}
We choose the fine-grained CARS196 and CUB200-2011, and the coarse-grained CIFAR10~\cite{krizhevsky2009learning} as the datasets for our deep metric learning experiments. We follow the conventional way to split the training and testing data:

(1) The CARS196 dataset~\cite{krause20133d} contains 16,185 images of 196 car models. The training set and testing set are composed of 8,144 images and 8,041 images, of 196 models.

(2) The CUB200-2011 dataset~\cite{wah2011caltech} includes 11,788 images of 200 bird species. The training set and testing set are composed of 5,994 images and 5,794 images, of 200 classes.

(3) The CIFAR10 dataset~\cite{krizhevsky2009learning} contains 60,000 32x32 color images of 10 classes. We randomly select 100 images per class as the testing set, then the rest 59,000 images as database set. From the database set, we randomly choose 500 images per class as the training set.

The experiment results of CARS196 and CUB200-2011 are reported on the testing set, and the results on CIFAR10 are reported by querying the testing set in the database set.
\subsubsection{Implementation Details and Evaluation Metrics}
Our method was implemented based on TensorFlow. We adopt the AlexNet~\cite{krizhevsky2012imagenet} for deep metric learning. In order to generate d-dimensional features $\bm{h}_i\in \mathbb {R}^M$, we replace the last classifier layer $fc8$ with an embedding layer of $M$ hidden units. For training, we fine-tune the layers except of the embedding layer from the model pre-trained on ImageNet and train the embedding layer, all through back-propagation. We use mini-batch stochastic gradient descent (SGD) with 0.9 momentum, and fix the mini-batch size of images as 100, except the relative N-pair methods on CIFAR10, which is set to 20 instead. All the input images of these experiments are resized into the 227 x 227 to fit the input size of AlexNet.

To evaluate the performance of different deep metric learning methods, we follow the protocol in~\cite{oh2016deep,wang2017deep} to conduct experiments on both clustering tasks and retrieval tasks. For the clustering tasks, we make experiment on CUB200-2011 and CARS196, and use NMI and F1 score to measure the performance of different methods. NMI is defined by the ratio of mutual information and the average entropy of clusters and the entropy of labels. F1 metric computes the harmonic mean of precision($P$) and recall($R$), and F1 = $\frac{2PR}{P+R}$. For image retrieval tasks, we calculate the Recall@K for the experiment results on CUB200-2011 and CARS196, and record the MAP and F1 metric for the experiment results on CIAFR10. Recall@K is computed by that each query will score 1 if an semantic similar image is retrieved in K nearest neighbors from test data. MAP is the mean of the Average Precision (AP), and AP of each query is computed as $\text{AP@}T=\frac{\sum_{t=1}^T P(t)\delta(t)}{\sum_{t'=1}^T \delta(t')}$, where $T$ is the number of top-returned images, $P(t)$ denotes the precision of top $t$ retrieved results, and $\delta(t)=1$ if the $t$-th retrieved result is true neighbor of the query, otherwise $\delta(t)=0$. We use MAP@59000 and F1@5000 as evaluation criteria for CIFAR10, where MAP@59000 means that MAP on the returned top-59000 images, and F1@5000 means F1 scores on the returned top-5000 images.
\vspace{-1em}
\begin{table*}[t]
\caption{\label{tab4}MAP@50000 of Hamming Ranking on CIFAR10 and NUS-WIDE with CNN-F. DPSH* denotes re-running the code provided by the authors of DPSH.}
\centering
\begin{tabular}{|c|c|c|c|c|c|c|c|c|c|}
\hline
 \multirow{2}{*}{method}&\multicolumn{4}{|c|}{CIFAR10}&{\multirow{2}{*}{method}}&\multicolumn{4}{|c|}{NUS-WIDE}\\
 \cline{2-5}\cline{7-10} &16 bits&24 bits&32 bits&48 bits&\multirow{2}{*}{}&16 bits&24 bits&32 bits&48 bits\\
 \hline
DSRH~\cite{zhao2015deep}&0.608&0.611&0.617&0.618&DSRH~\cite{zhao2015deep}&0.609&0.618&0.621&0.631\\
\hline
DSCH~\cite{zhang2015bit}&0.609&0.613&0.617&0.620&DSCH~\cite{zhang2015bit}&0.592&0.597&0.611&0.609\\
\hline
DRSCH~\cite{zhang2015bit}&0.615&0.622&0.629&0.631&DRSCH~\cite{zhang2015bit}&0.618&0.622&0.623&0.628\\
\hline
DTSH~\cite{wang2016deep}&0.915&0.923&0.925&0.926&DTSH~\cite{wang2016deep}&0.756&0.776&0.785&0.799\\
\hline
DPSH*~\cite{li2015feature}&0.903&0.885&0.915&0.911&DPSH~\cite{li2015feature}&0.715&0.722&0.736&0.741\\
\hline
DSNRH(Ours)&\textbf{0.925}&\textbf{0.932}&\textbf{0.934}&\textbf{0.940}&DSNRH(Ours)&\textbf{0.830}&\textbf{0.840}&\textbf{0.852}&\textbf{0.862}
\\
 \hline
\end{tabular}
\end{table*}

\subsubsection{Results and Analysis}

Table~\ref{tab1} and Table~\ref{tab2} show the performance of deep metric learning methods on CARS196 and CUB200-2011, and we obtain the results by comparing the Euclidean-based deep metric learning methods with our DSML under various embedding sizes, including 16, 32, 64. We observe that the proposed SNR-based metric boosts the performance of state-of-the-art metric learning approaches on all the benchmark datasets. The experiment results on CARS196 and CUB200-2011 datasets show similar tendency: combined with our DSML, the performance improvements on contrastive, triplet, lifted, N-pair loss are all significant.

Figure~\ref{ff2} shows the retrieval results of Recall@K on CARS196 and CUB200-2011, at the embedding size of 64. The results show that our DSML obviously outperforms other corresponding Euclidean-based methods. We can find that the most prominent curve in Figure~\ref{ff2} is DSML(tri), which have the highest performance over other methods.

\begin{figure}[t]
\centering
\includegraphics[width=0.5\textwidth]{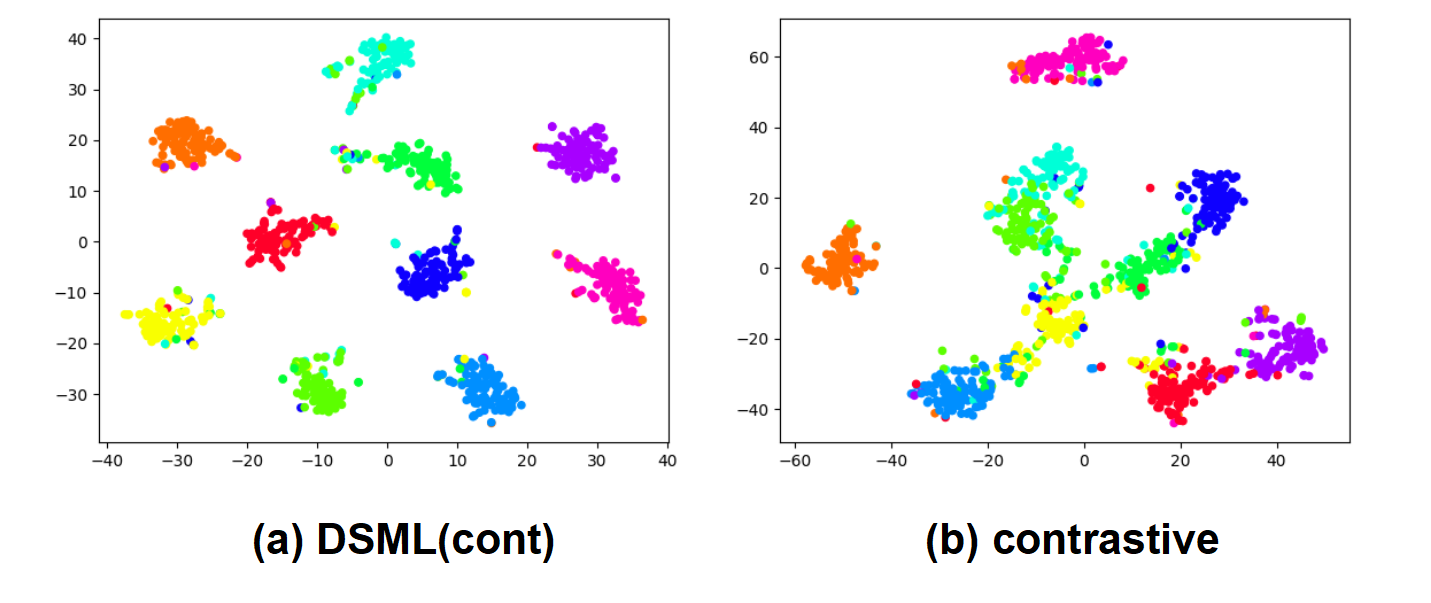}
\caption{\label{ff3} The t-SNE visualization of the features learned by our DSML(cont) method and the contrastive method with Euclidean distance on CIFAR-10 dataset}
\vspace{-1em}
\end{figure}

Table~\ref{tab3} shows the comparative results of retrieval tasks on CIFAR10 dataset with two retrieval strategies: Euclidean ranking and Hamming ranking. Euclidean ranking is the general retrieval approach, which computes the Euclidean distance of real-valued features to generate the rank list. Hamming ranking is on the basis of the binary features and computes the Hamming distance. To obtain the binary codes, in our experiment, we make a quantization on real-valued embedding by sign function. As shown the Table~\ref{tab3}, our DSML method still has superior results than the related Euclidean distance based metric learning methods. The unsatisfactory results on lifted loss and N-pair loss indicate that these losses are not suitable for the CIFAR10 dataset with a large number of images but only ten classes.

Figure~\ref{ff3} shows the t-SNE visualizations~\cite{maaten2008visualizing} of the features learned by DSML(cont) and contrastive on CIFAR-10. The result indicates that the features learned by our DSML(cont) exhibit more clear discriminative structures, while the original contrastive loss presents relative vague structures.

The encouraging performances of our DSML is because our SNR distance metric has more power to enlarge the inter-class distances and reduce the intra-class distances than the traditional Euclidean distance metric. Besides, our SNR distance metric can also preserve correlation information in image pairs to improve the performance in learned embeddings.

\subsection{Experiments on Hashing Learning}


\subsubsection{Datasets}
We evaluate the performance on two datasets: CIFAR10 and NUS-WIDE, and the results are reported by querying the testing set in the database set.

(1) For CIFAR10~\cite{krizhevsky2009learning}, we randomly select 1000 images per class as the test query set, and the rest images are selected as the training set and database set.

(2) NUS-WIDE~\cite{chua2009nus} is consisted of 269,648 images associated with 81 tags. Similar to DPSH~\cite{li2015feature} and DTSH~\cite{wang2016deep}, we utilize 21 most frequent concepts to select 195,834 images as experimental dataset. We randomly sample 100 images in each class (2,100 images in total) as the test query images, and the remaining images are used as the training set and database set.
\subsubsection{Implementation Details and Evaluation Metrics}
Similar to DPSH~\cite{li2015feature} and DTSH~\cite{wang2016deep}, we deploy the CNN-F network architecture in our DSNRH. The input images of our experiments are resized into the 224 x 224. We also use mini-batch stochastic gradient descent (SGD) with 0.9 momentum, and give the mini-batch size of images as 100.

We report MAP@50000 results based on the top 50,000 returned neighbors, at the binary codes length of 16, 24, 32, and 48 bits. In order to have a fair comparison, most of the existing experiment results are directly reported from previous works.
\subsubsection{Results and Analysis}
We compare the retrieval performance of our DSNRH with five deep hashing methods, including DPSH~\cite{li2015feature}, DTSH~\cite{wang2016deep}, DRSCH~\cite{zhang2015bit}, DSCH~\cite{zhang2015bit}, DSRH~\cite{zhao2015deep}. The MAP results of our experiment are presented in Table~\ref{tab4}. We can find that our DSNRH substantially outperforms all the other methods. The performance of some deep hashing methods, including DSRH, DSCH and DRSCH, are inferior to our method, and their average MAP results are only above 60\% in two datasets. DPSH and DTSH are also based on the CNN-F network architecture, but they have lower precision. The outstand performance of our DSNRH demonstrates that our SNR-based metric can also improve the robustness of hashing code learning.

\section{Conclusion}
In this paper, we propose a robust distance metric based on Signal-to-Noise Ratio~(SNR) as similarity measurement for deep metric learning. By replacing the Euclidean distance measurement with our SNR distance metric, we construct deep SNR-based metric learning, which can generate more discriminative features than the Euclidean-based deep metric learning. In the extensive experiments for image clustering and retrieval tasks, our DSML has shown its superiority to the state-of-the-art deep metric learning methods on three benchmarks. As an extension of our SNR-based metric, we also propose a deep SNR-based hashing method, and the experiments on two benchmarks show the outstanding performance of DSNRH. Based on the generality of our SNR-based similarity metric, we believe our SNR-based metric is promising to be further applied to more deep learning models.

\section*{Acknowledgement}
This work was partially supported by the National Natural Science Foundation of China under Grant Nos.~61871052, 61573068, 61471048, and 61375031, and by the Beijing Nova Program under Grant No.~Z161100004916088.

{\small
\bibliographystyle{ieee_fullname}
\bibliography{aaai}
}

\end{document}